%% file: main.tex
\documentclass[10pt,twocolumn,letterpaper]{article}

\usepackage{iccv}              % To produce the CAMERA-READY version
\usepackage{times}
\usepackage{epsfig}
\usepackage{graphicx}
\usepackage{amsmath}
\usepackage{array}
\usepackage{amssymb}
\usepackage{subcaption}
\usepackage{multirow}
\usepackage{subcaption}
\usepackage{indentfirst}
\usepackage{verbatim}
\usepackage{bm}
\usepackage{kantlipsum}
\usepackage{hhline}
\usepackage{color}
\usepackage{booktabs}
\usepackage{bbding}
\usepackage{makecell} 
\usepackage{colortbl} % Add this to your preamble for table coloring
\usepackage{xcolor}

\definecolor{iccvblue}{rgb}{0.21,0.49,0.74}
\usepackage[pagebackref,breaklinks,colorlinks,allcolors=iccvblue]{hyperref}

\def\paperID{5131} % *** Enter the Paper ID here
\def\confName{ICCV}
\def\confYear{2025}

\title{RainbowPrompt: Diversity-Enhanced Prompt-Evolving for Continual Learning}

\author{Kiseong Hong\\
Department of AI\\
Chung-Ang University\\
{\tt ghdrltjd@cau.ac.kr}
\and
Gyeong-hyeon Kim\\
School of CSE\\
Chung-Ang University\\
{\tt leonardkkh@cau.ac.kr}
\and
Eunwoo Kim$\thanks{Corresponding author.}$\\
School of CSE\\
Chung-Ang University\\
{\tt eunwoo@cau.ac.kr}
}

\begin{document}
\maketitle

%%%%%%%%% ABSTRACT
\begin{abstract}
Prompt-based continual learning provides a rehearsal-free solution by tuning small sets of parameters while keeping pre-trained models frozen.
To meet the complex demands of sequential tasks, it is crucial to integrate task-specific knowledge within prompts effectively. 
However, existing works rely on either fixed learned prompts (i.e., prompts whose representations remain unchanged during new task learning) or on prompts generated from an entangled task-shared space, limiting the representational diversity of the integrated prompt.
To address this issue, we propose a novel prompt-evolving mechanism to adaptively aggregate base prompts (i.e., task-specific prompts) into a unified prompt while ensuring diversity.
By transforming and aligning base prompts, both previously learned and newly introduced, our approach continuously evolves accumulated knowledge to facilitate learning new tasks.
We further introduce a learnable probabilistic gate that adaptively determines which layers to activate during the evolution process.
We validate our method on image classification and video action recognition tasks in class-incremental learning, achieving average gains of 9.07\% and 7.40\% over existing methods across all scenarios.
\end{abstract}

%%%%%%%%% BODY TEXT
\section{Introduction}
Continual learning (CL) aims to enable a model to learn sequential tasks while retaining knowledge from previous tasks. 
However, sequential learning faces a major challenge, catastrophic forgetting~\cite{french1999catastrophic}, where learning new tasks deteriorates performance on previously learned tasks.
To address this, various CL approaches~\cite{jung2020continual,saha2021gradient,jin2023growing} have been proposed.
A recent paradigm in CL, known as prompt-based CL (PCL)~\cite{wang2022learning,qiao2023prompt}, has shown promise in mitigating catastrophic forgetting with its rehearsal-free nature, which applied to a pre-trained vision transformer (ViT)~\cite{dosovitskiy2020image}.
These methods leverage prompt-tuning~\cite{lester2021power}, a transfer learning strategy initially developed for natural language processing (NLP).
Instead of modifying the weights of ViT, prompt-tuning keeps the model frozen and fine-tunes additional small sets of parameters called prompts. 
These prompts act as task-specific instructions, guiding the model without the need for direct weight updates.

\begin{figure}[t]
\centering
\begin{subfigure}[b]{0.56\columnwidth}
    \includegraphics[width=\textwidth]{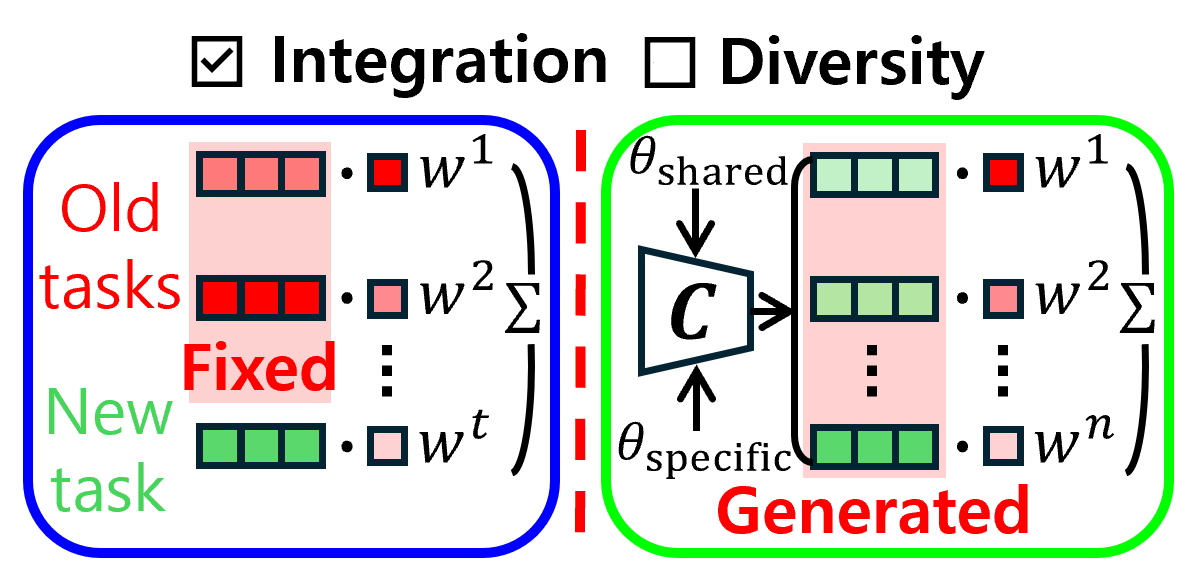}
    \caption{Existing prompt integration}
    \label{fig:existing_inte}
\end{subfigure}
\hfill
\begin{subfigure}[b]{0.43\columnwidth}
    \includegraphics[width=\textwidth]{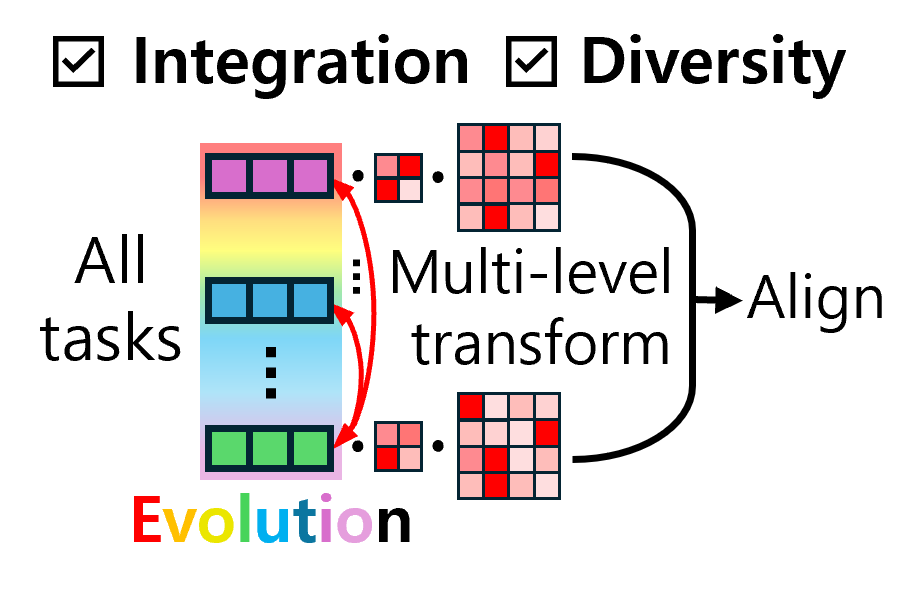}
    \caption{Prompt-evolution (ours)}
    \label{fig:proposed_inte}
\end{subfigure}
\caption{
Conceptual illustration of (a) existing prompt integration approaches and (b) the proposed prompt-evolving approach. 
(a) Existing approaches integrate prompts using input-conditioned weights $w$, with fixed old task prompts (blue box) or those generated between task-shared and task-specific spaces (green box).
(b) Our approach progressively transforms and aligns prompts to make their representations more adaptable to new tasks.}
\label{fig:concept}
\end{figure}

Existing PCL works either select task-relevant prompts from a prompt pool~\cite{wang2022learning} or extend this approach by incorporating a task-invariant prompt shared across all tasks~\cite{wang2022dualprompt}.
Recent studies attempt to integrate prompts into a unified prompt~\cite{smith2023coda,roy2024convolutional}. 
They produce a prompt by combining prompts via an input-conditioned weighted sum~\cite{smith2023coda} or by generating task-specific prompts and merging them~\cite{roy2024convolutional} (see Fig. \ref{fig:existing_inte}).
However, they often yield prompts with limited representational diversity. 
This limitation arises from relying on learned representations that remain unchanged without adapting to new tasks~\cite{smith2023coda} or deriving prompts in a task-shared space vulnerable to task interference and dominance~\cite{roy2024convolutional}.
This reduces adaptability and generalization, highlighting the need for more effective strategies to balance integration with diversity.

\begin{figure*}[t]
\begin{center}
\begin{subfigure}[b]{0.32\textwidth}
    \includegraphics[width=\textwidth]{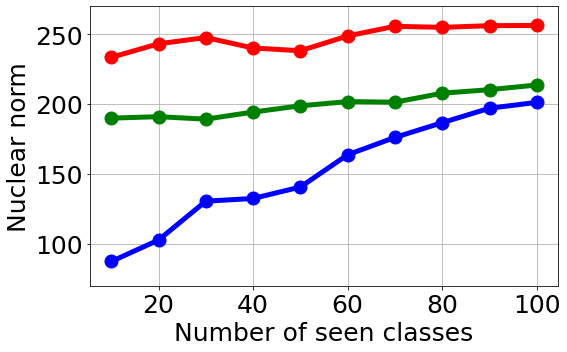}
    \caption{Number of seen classes vs. Diversity}
    \label{fig:nuclear_a}
\end{subfigure}
\hfill
\begin{subfigure}[b]{0.32\textwidth}
    \includegraphics[width=\textwidth]{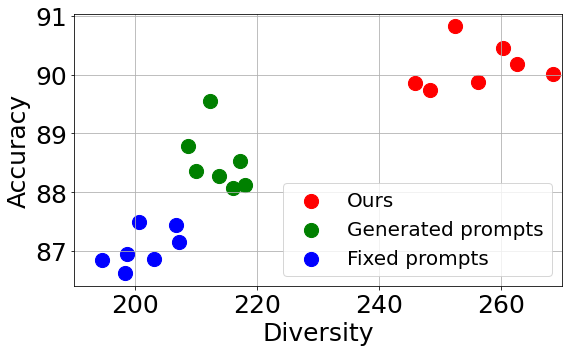}
    \caption{Diversity vs. Accuracy}
    \label{fig:nuclear_b}
\end{subfigure}
\hfill
\begin{subfigure}[b]{0.32\textwidth}
    \includegraphics[width=\textwidth]{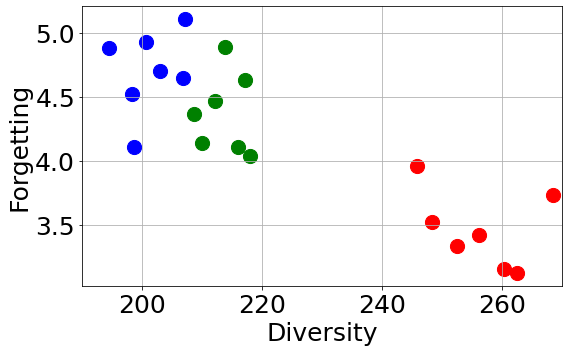}
    \caption{Diversity vs. Forgetting}
    \label{fig:nuclear_c}
\end{subfigure}
\end{center}
%\vspace*{-2mm}
\caption{Results of the representation diversity in a class-incremental learning scenario with a 10-task setting on CIFAR-100.
We compare ours with two baselines using fixed~\cite{smith2023coda} and generated~\cite{roy2024convolutional} prompts.
In (a), we measure the average nuclear norm of the prompt at each time step as new classes are introduced, using test samples from all seen classes.
We insert prompts into all layers and calculate diversity using the prompt in the last layer.
In (b) and (c), we compare the representation diversity (after all classes are learned) against accuracy and forgetting, respectively, with each dot denoting a distinct random trial.
}
\label{fig:singular}
\end{figure*}

Furthermore, our findings in Fig. \ref{fig:singular} reveal that this limitation hinders achieving high accuracy while minimizing forgetting during sequential task learning.
We observe that existing methods fail to achieve optimal knowledge integration with enhanced diversity\footnote{We analyze the nuclear norm of the prompt as a measure of representation diversity. 
A higher nuclear norm reflects greater representation diversity and improves the discriminability between classes~\cite{han2021rethinking,cui2020towards}.}.

Merging prompts via a weighted sum without adequate adaptation to new tasks yields low representational diversity (blue line in Fig.~\ref{fig:nuclear_a}).
This limitation restricts the ability to capture distinctive features introduced by new classes.
Generating prompts for new classes improves representational diversity, as it adapts to class-specific traits (green line in Fig. \ref{fig:nuclear_a}).
However, the diversity of prompts generated from the increasingly entangled task-shared space remains insufficient, as overlapping knowledge across tasks leads to less distinctive representations.
Figures \ref{fig:nuclear_b} and \ref{fig:nuclear_c} demonstrate that higher representation diversity improves accuracy and reduces forgetting, respectively.
These analyses underscore the importance of optimal knowledge integration to ensure diversity and preserve previously learned knowledge in sequential task learning.

In this work, we propose a prompt-evolving approach to integrate task-specific knowledge into a diversity-enhanced representation. 
Prompts undergo transformation and alignment, gradually adapting their representations to new tasks while preserving previously learned information.
Our method progressively reconfigures the prompts to seamlessly integrate complementary knowledge from both previous and current tasks.
It consists of two steps: attention-based transformation and task-guided alignment.
%Our attention-based transformation self-identifies influence patterns for new tasks across multiple prompts. 
Our attention-based transformation assesses the relevance of multiple prompts to new tasks.
Unlike existing works that derive prompt weights via cosine similarity between input-conditioned queries and prompt keys~\cite{smith2023coda}, our method dynamically reweights the contribution of each prompt based on its relevance to new tasks at multi-level granularity.
The task-guided alignment further refines these representations by progressively tailoring them to a new task.
It incorporates non-linear transformations to align the representations of the new task while preserving the intrinsic properties of each prompt.
Finally, we introduce RainbowPrompt, a novel prompt that integrates enhanced representations of task-specific prompts.
We further optimize the proposed method by introducing a learnable probabilistic gate that adaptively regulates layer activation during evolution, leveraging task-specific differences.

To demonstrate the effectiveness of RainbowPrompt, we compare it with existing prompt-based CL approaches~\cite{wang2022learning,wang2022dualprompt,smith2023coda,roy2024convolutional,qiao2023prompt,gao2024consistent} in class-incremental learning scenarios.
The evaluation includes image classification tasks using ImageNet-R~\cite{hendrycks2021many}, CIFAR-100~\cite{krizhevsky2009learning}, and CUBS~\cite{wah2011caltech}. 
We also evaluate video action recognition tasks on UCF-101~\cite{soomro2012ucf101} and ActivityNet~\cite{caba2015activitynet}.
Extensive experiments show that the proposed method significantly outperforms the compared methods across all benchmarks, demonstrating its effectiveness and versatility in image and video tasks.
The main contributions of this work are as follows: 
\begin{itemize}
    \item 
    We propose a novel prompt-evolving approach for CL, enabling effective knowledge integration while ensuring diversity across tasks.
    
    \item 
    Our method employs attention-based transformation and task-guided alignment to integrate knowledge from previous tasks while adapting to new tasks.
    
    \item Experimental results on a wide range of tasks demonstrate that ours outperforms its competitors with an overall average margin of 8.23\% across all scenarios.
\end{itemize}

\section{Related Work}
\noindent \textbf{Continual Learning.}
Continual learning (CL) aims to progressively acquire new knowledge while retaining previously learned information~\cite{de2021continual,hong2025exploration}. 
A major challenge in CL is catastrophic forgetting~\cite{french1999catastrophic}, where learning new tasks degrades performance on earlier ones. 
Solutions to this challenge can be broadly divided into three categories: 
Regularization-based methods~\cite{jung2020continual,aljundi2018memory} mitigate catastrophic forgetting by adding penalty terms that consolidate knowledge from earlier tasks.
Rehearsal-based methods tackle catastrophic forgetting by retaining a subset of the data from previous tasks~\cite{rebuffi2017icarl,lopez2017gradient,chaudhry2018efficient} or generating data for previous tasks using adversarial techniques~\cite{shin2017continual,wu2018memory,ostapenko2019learning}. 
Architecture-based methods~\cite{yoon2017lifelong,hung2019compacting,yan2021dynamically,jin2022helpful} dynamically adjust the network structure to accommodate new tasks. 
Despite the advancements of these approaches, the emergence of foundation models trained on large-scale data has driven a paradigm shift in CL~\cite{wang2022learning}, moving towards replay-free and parameter-efficient approaches \cite{jung2023generating, liang2024inflora}.

\noindent \textbf{Prompt-based Continual Learning.}
Prompting~\cite{liu2023pre} initially involved manually designed task-specific instructions to induce desired responses from pre-trained models.
In the context of CL, prompt-based methods~\cite{wang2022learning,qiao2023prompt} retain task-specific knowledge without rehearsal buffers. 
They use prompts as keys to retrieve stored knowledge, eliminating the need for model parameter updates.
L2P~\cite{wang2022learning} introduces prompt-tuning with key-query matching to retrieve task-relevant knowledge. DualPrompt~\cite{wang2022dualprompt} further introduces task-invariant prompts along with task-specific prompts to distinctly encode task-relevant and task-agnostic knowledge.
CODA-Prompt~\cite{smith2023coda} introduces a decomposed attention-based prompting method, while ConvPrompt~\cite{roy2024convolutional} leverages convolution for prompt generation.
PGP constrains prompt updates to be orthogonal to prior prompt directions via a projection mechanism.
CPrompt \cite{gao2024consistent} introduces consistency prompting to address inconsistencies between the classifier and the prompt.

\noindent \textbf{Limitations.}
Among the competitive approaches, CODA-Prompt~\cite{smith2023coda} and ConvPrompt~\cite{roy2024convolutional} focus on integrating multiple base prompts but often overlook diversity within the resulting unified prompt.
Fixed representations hinder the capture of diverse information~\cite{chen2018lifelong,cha2021co2l}, while generated representations shaped by a task-shared space often suffer from task interference and dominance, resulting in overfitting and limited generalization~\cite{ma2023understanding,khattak2023self}.
In contrast, we address these limitations by progressively evolving the representations of all accumulated base prompts.
We accumulate independent base prompts for each task and integrate them into a unified prompt for optimal knowledge integration.

\begin{figure*}[t]
\begin{center}
\includegraphics[width=\textwidth]{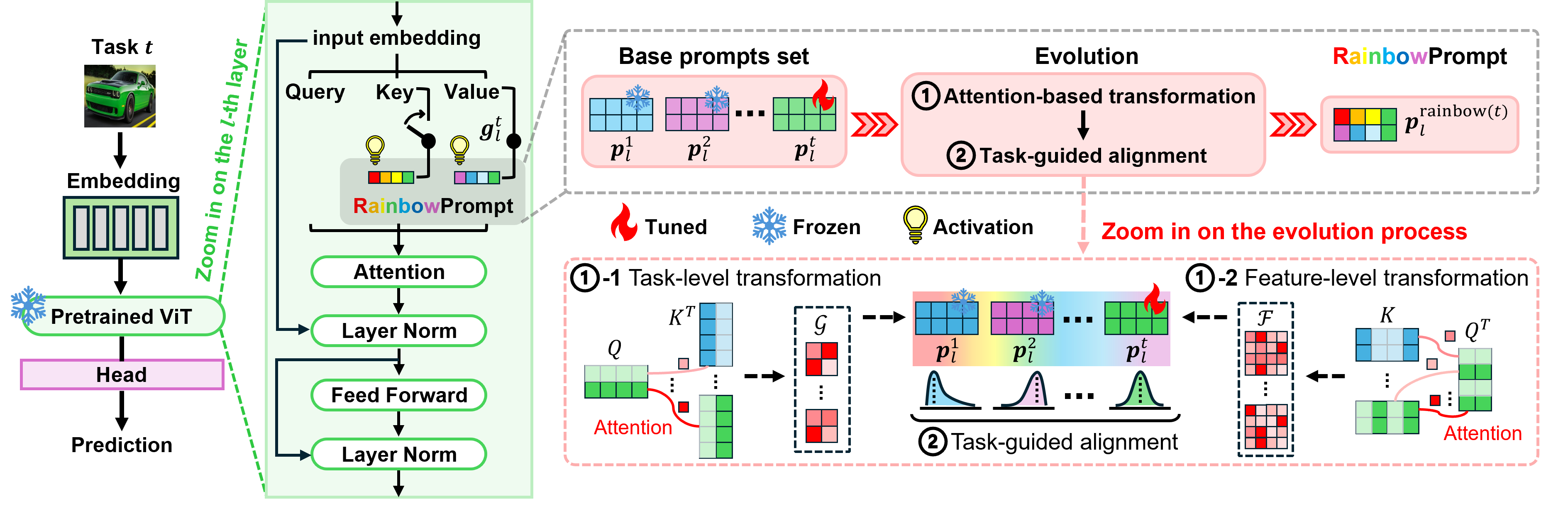}
\end{center}
%\vspace*{-2mm}
\caption{Illustration of the proposed framework.
%When a new task $t$ arrives, {\color{red}the base prompt set, containing all task-specific prompts up to the $t$-th time step, is evolved} through attention-based transformation and task-guided alignment. 
When a new task $t$ arrives, we evolve the base prompt representations, comprising all task-specific prompts up to the $t$-th time step, through attention-based transformation and task-guided alignment.
Finally, RainbowPrompt, $\bm{p}_l^{\text{rainbow}(t)}$, is constructed at each layer $l$ by integrating diverse knowledge from the accumulated base prompts.
We further introduce a learnable probabilistic gate $\bm{g}_{l}^{t}$ to selectively insert $\bm{p}_l^{\text{rainbow}(t)}$ into intermediate layers.
During testing, the proposed method requires the RainbowPrompts $\{\bm{p}_l^{\text{rainbow}(i)}\}_{l=1}^{L_i}$ for each task $i$, which are appended to $L_i$ selected layers determined by $\{\bm{g}_{l}^{i}\}_{l=1}^{L}$.
}
\label{fig:2}
\end{figure*}

\section{Methodology}
\subsection{Prerequisites}
Continual learning (CL) enables a model to progressively learn knowledge from a sequence of tasks $\{\mathcal{T}^1, \mathcal{T}^2, \dots, \mathcal{T}^T\}$.
The $t$-th task $\mathcal{T}^t$ consists of $N_t$ input-label pairs $\{(\bm{x}_i, y_i)\}_{i=1}^{N_t}$, where $\bm{x}_i \in \mathcal{X}^{t}$ represents the input data and $y_i \in \mathcal{Y}^t$ denotes the corresponding label. 
In this work, we focus on class-incremental learning, where task identities are unknown during testing \cite{hsu2018re}. 
We also adopt a practical rehearsal-free setting \cite{wang2022dualprompt}, which prohibits the storage of data from previous tasks.

In prompt-based continual learning (PCL), the model $f_{\theta}(\cdot)$ is typically chosen to be a pre-trained vision transformer (ViT) \cite{dosovitskiy2020image}.
Rather than adjusting the weights of ViT directly, we use small learnable parameters, i.e., prompts $\bm{p} \in \mathbb{R}^{L_p \times D}$, to provide task-specific instructions \cite{lester2021power}, where $L_p$ represents the prompt length and $D$ is the embedding dimension.
Specifically, we employ prefix tuning ($\text{P-T}$) \cite{li2021prefix}, where the prompts comprising $\bm{p}_K \in \mathbb{R}^{L_p/2 \times D}$ and $\bm{p}_V \in \mathbb{R}^{L_p/2 \times D}$ are prepended to the key and value representations in the multi-head self-attention layers of ViT.

\subsection{Prompt-Evolving Mechanism}
Our goal is to consolidate task-specific prompts into a cohesive prompt through a prompt-evolving mechanism, which dynamically reconfigures their representations to adapt to new tasks while enhancing representational diversity.
At each time step $t$, a new task is introduced, with its corresponding base prompts (i.e., task-specific prompts) $\bm{p}^{t} = \{\bm{p}_{l}^{t}\}^{L}_{l=1}$, where $\bm{p}_{l}^{t} \in \mathbb{R}^{L_p \times D}$ is the base prompt at layer $l$, and $L$ is the number of layers.
To facilitate sequential learning, we maintain a set of accumulated prompts denoted as $\bm{\mathcal{P}} = \{\bm{p}^{1}, \bm{p}^{2}, \dots, \bm{p}^{t}\}$, which includes all base prompts up to time step $t$.
Our approach seeks to integrate these prompts into unified prompts, i.e., $\bm{p}^{\text{rainbow}(t)} = \{\bm{p}^{\text{rainbow}(t)}_l\}_{l=1}^{L}$, where $\bm{p}^{\text{rainbow}(t)}_l$ is the unified prompt at layer $l$, without hurting the base prompts.

The accumulated prompts $\bm{\mathcal{P}}$ present a challenge when combining them in continual learning. 
Direct integration of prompts risks losing knowledge from previously learned tasks, while freezing them restricts adaptation and hinders knowledge transfer to the $t$-th task.
To address this, we update the base prompts for the current task at each time step while keeping the base prompts from previous tasks frozen to preserve their knowledge. 
To facilitate knowledge transfer to new tasks, we transform and align the representations of all base prompts by introducing learnable components.
Specifically, we apply self-attention to dynamically reweight the contributions of accumulated prompts, facilitating their transformation by emphasizing salient information while suppressing less relevant details. 
This adaptive reweighting mitigates knowledge dilution and enhances representational diversity.
By aligning the prompts to their transformed representations for the new task, we encapsulate the diverse knowledge in $\bm{\mathcal{P}}$ into a set of unified prompts.
The overall framework is illustrated in Fig. \ref{fig:2}.

\subsubsection{Attention-based Transformation}
\label{sec: RA}
We first apply an attention-based transformation that enables accumulated prompts to self-identify their influence on the new task via two-level interactions: task-level and feature-level.
Interactions among prompt vectors capture overall contributions at the task level, while interactions among the individual features (elements) of each vector measure fine-grained contributions.
Before the transformation process, we precede a task-conditioning step to inject task-relevant information into $\bm{\mathcal{P}}_l$\footnote{We describe the method at a specific layer $l$ for notational convenience.}.

We use a learnable task embedding vector $\bm{e}^t \in \mathbb{R}^D$ to compute attention weights that emphasize the task-relevant components of $\bm{\mathcal{P}}_l$ at the $t$-th time step. 
We condition the set of base prompts by $\bm{\mathcal{P}}_l \leftarrow \text{softmax}\bigl(\sigma(\bm{e}^t)\bm{\mathcal{P}}_l^T / \sqrt{d_p} \bigr) \bm{\mathcal{P}}_l$,  
where $\sigma(\cdot)$ and $d_p$ denote the broadcasting function and the dimensionality of the $\bm{\mathcal{P}}_l$, respectively.

We treat $\bm{p}^{\text{new}}_{l} = \bm{p}^{t}_{l}$ as the query and use the concatenated task-conditioned base prompts set $\bm{\mathcal{P}}_l$ as the key and value inputs to the transformation process.
First, we project each prompt into a lower-dimensional space to align their dimensions for the transformation, reducing the complexity of subsequent operations:
\begin{equation}
\label{eq:projection}
    Q = W^Q_l \bm{p}^{\text{new}}_l,\; K = W^K_l \bm{\mathcal{P}}_l,\; V = W^V_l \bm{\mathcal{P}}_l,
\end{equation}
where $W^{Q}_{l} \in \mathbb{R}^{D \times D_p}$, $W^{K}_{l} \in \mathbb{R}^{D \times D_p}$, and $W^{V}_{l} \in \mathbb{R}^{D \times D_p}$ are learnable projection matrices for the respective query, key, and value at the $l$-th layer, and $D_p$ represents the projected dimension ($D_p \ll D$).

For the task-level transformation, we define the attention-based transformation function, $\text{AT}(\cdot)$, which computes the inter-task affinity matrix $\mathcal{G}$ from the attention between $Q$ and $K$, and then uses $\mathcal{G}$ to weight the value representations:
\begin{equation}
    \text{AT}(Q, K, V) = \mathcal{G} \cdot V \triangleq \tilde{V} \in \mathbb{R}^{t \times L_p \times D_p}, 
\end{equation}
where $\mathcal{G} = \text{softmax}\left( Q K^T / \sqrt{d_k} \right)$, $d_k$ denotes the dimensionality of the key, $\cdot$ represents the matrix multiplication operation, and $\tilde{V}$ is the transformed representation.
$\mathcal{G}$ quantifies how much information from existing tasks should contribute to the new task, allowing us to adaptively weigh the influence of each task.
By integrating $\mathcal{G}$ with $V$, we transform the representations associated with the prompts of each task based on their contribution to the new task $t$.

The feature-level transformation captures cross-feature influences at a finer granularity, complementing the task-level transformation.
It leverages transposed query $Q^T$ and key $K$ to enable comparison of feature dimensions.
This approach is inspired by bilinear pooling~\cite{gao2016compact}, which enjoys multiplicative interactions between feature dimensions.
We update the output of the task-level transformation $\tilde{V}$ as
\begin{equation}
\text{AT}(Q, K, \tilde{V}) = \mathcal{F} \cdot \tilde{V}^T \triangleq \hat{V}\in \mathbb{R}^{t \times D_p \times L_p}, \end{equation}
where the inter-feature affinity matrix $\mathcal{F}$ is computed as $\mathcal{F} = \text{softmax}\left( Q^T K / \sqrt{d_k} \right)$.
It quantifies the contributions between individual features across tasks, capturing fine-grained dependencies. 
By applying $\mathcal{F}$ to $\tilde{V}$, the feature-level transformation integrates cross-feature influences, refining $\tilde{V}$ to incorporate contributions distinct from those of the task-level transformation.

To integrate the transformed representations, we redefine the output of the feature-level transformation as $\hat{V} \leftarrow \text{LN}(\bm{\mathcal{P}}_l + (\hat{V}^T W^{O}_{l})) \in \mathbb{R}^{t \times L_p \times D}$.
$\text{LN}(\cdot)$ represents layer normalization, and $W^{O}_{l} \in \mathbb{R}^{D_p \times D}$ is a learnable projection matrix. 
This integrates $\bm{\mathcal{P}}_l$ with the transformed representations, preserving the original information in $\bm{\mathcal{P}}_l$ while enriching the resulting representations with newly adapted features.

\subsubsection{Task-Guided Alignment}
\label{sec: PR}
Following the attention-based transformation, the task-guided alignment refines the transformed representations $\hat{V}$.
This decodes the representations to meet the traits of the new task, preserving the attributes of each prompt:
\begin{equation} 
\bm{\tilde{\mathcal{P}}}_l = \text{LN}(\hat{V} + \text{LT}(\hat{V})) \in \mathbb{R}^{t \times L_p \times D},
\end{equation} 
where $\text{LT}(x) = \text{max}(0, x W_l^1) W_l^2$ aligns the transformed representations to identify task-relevant patterns.
It refines their distribution across tasks within a reduced space $D_n$ ($D_n \ll D$)~\cite{meng2022locating,geva2020transformer}, with
$W_l^1 \in \mathbb{R}^{D \times D_n}$ and $W_l^2 \in \mathbb{R}^{D_n \times D}$ as learnable weight matrices (bias terms omitted for simplicity).

Finally, we derive a RainbowPrompt, $\bm{p}^{\text{rainbow}(t)}_{l}$, by combining the evolved representations of prompts from all tasks learned up to time step $t$ into a unified prompt: \begin{equation} \bm{p}^{\text{rainbow}(t)}_{l} = \frac{1}{t} \sum_{i=1}^{t} \bm{\tilde{\mathcal{P}}}_l[i] \in \mathbb{R}^{L_p \times D}, \end{equation} 
where $\bm{\tilde{\mathcal{P}}}_l[i]$ denotes the evolved representation of the $i$-th base prompt.
$\bm{p}^{\text{rainbow}(t)}_{l}$ is a distinct type of prompt that differs from task-specific and task-shared prompts by evolving and integrating accumulated knowledge without predefined roles.
Note that the proposed prompt-evolving mechanism is performed exclusively during training, where task-specific RainbowPrompts are produced and stored. 
At test time, they are directly used for prediction without requiring the evolution components.

\subsection{Adaptive Prompting}
\label{sec: ap and optimization} 
Determining which layers to insert RainbowPrompts poses an additional challenge due to varying learning complexities across tasks.
Manually selecting these layers is impractical, as it fails to account for task-specific differences \cite{hong2025exploration}.
To this end, we introduce a task-specific learnable probabilistic gate $\bm{G}^t = \{\bm{g}_{l}^{t}\}^{L}_{l=1}$ to learn where to insert RainbowPrompts in the model.
$\bm{g}_{l}^{t}$ is a Bernoulli random variable deciding whether to insert a RainbowPrompt at layer $l$.
Since the discrete nature of $\bm{g}_{l}^{t}$ prevents gradient-based optimization, we employ the Gumbel-Softmax trick~\cite{jang2016categorical} to relax it into a differentiable form: 
\begin{equation} \hat{g}_{l}^{t}(p) = \frac{\exp\left(\log \delta_{l}^{t}(p) + Z_{l}^{t}(p) / \tau \right)}{\sum_{i \in \{0,1\}} \exp\left(\log \delta_{l}^{t}(i) + Z_{l}^{t}(i) / \tau \right)},
\end{equation}
where $p \in \{0,1\}$, $Z_{l}^{t} = -\log(-\log U_{l}^{t})$ is the Gumbel noise generated from $U_{l}^{t} \sim \mathcal{U}[0,1]$, and $\tau$ controls the temperature for continuous relaxation. 
We sample discrete task-specific decisions from the learned distribution $\delta_{l}^{t} = [\alpha_{l}^{t},\, 1-\alpha_{l}^{t}]$, where $\alpha_{l}^{t}$ is the probability of prompt insertion for layer $l$.
This flexible activation aligns more effectively with adaptive prompting, compared to the manual prompting~\cite{smith2023coda,roy2024convolutional}.
Section \ref{sec: adaptive} provides a detailed discussion on this aspect.

\subsection{Optimization}
We jointly optimize $\delta_{l}^{t}$, the new task prompt $\bm{p}^{t}_{l}$, and the learnable parameters of the prompt-evolving mechanism, defined as $W^{\text{evolution}} = \{(W^{Q}_{l}, W^{K}_{l}, W^{V}_{l}, W^{O}_{l}, W^{1}_{l}, W^{2}_{l}) \}^{L}_{l=1}$, using the cross-entropy loss.
To encourage a sparse yet effective RainbowPrompt insertion pattern, we impose a regularization term to suppress insertion probabilities $\alpha_{l}^{t}$, i.e., $\mathcal{L}_{\text{sparse}} = \sum_{l \leq L} \log \alpha_{l}^{t}$.
In addition, we optimize the task embedding vector $\bm{e}^t$ using a matching loss~\cite{wang2022dualprompt}, $\mathcal{L}_{\text{match}} = \gamma(q(\bm{x}), \bm{e}^t)$, where $\bm{x} \in \mathcal{T}^t$, $\gamma$ denotes cosine similarity, and \(q(\cdot)\) is a query function~\cite{wang2022dualprompt}.
Finally, the proposed method minimizes the following total loss:
\begin{equation} \min_{\Theta^t} \sum_{i=1} \text{CE}(\bm{z}_i, y_i) + \lambda_\text{s} \mathcal{L}_{\text{sparse}} +  \lambda_\text{m} \mathcal{L}_{\text{match}},
\end{equation}
where $\Theta^t = \{\bm{p}^{t}, \bm{e}^t, \bm{G}^t, W^{\text{evolution}}, \phi \}$, where $\phi$ denotes a classifier.
\text{CE} is the cross-entropy loss, $\bm{z}_i$ represents the final output, and $\lambda_\text{s}$ and $\lambda_\text{m}$ are balancing parameters, both set to 0.01 in our experiments.

\begin{table*}[t]
%\vspace*{-5mm}
\caption{
Results on ImageNet-R and CIFAR-100.
$A_{N}$ and $F_{N}$ represent the average accuracy and forgetting for the $N$-task setting, respectively.
The best performance is highlighted in bold.}
\label{fig:combined}
\centering
\resizebox{\linewidth}{!}{%
\begin{tabular}{l||cccc||cccc}
\Xhline{1.pt} 
\multirow{2}{*}{\textbf{Method}} 
& \multicolumn{4}{c}{\textbf{ImageNet-R}} & \multicolumn{4}{c}{\textbf{CIFAR-100}} \\ \cline{2-9} 
& \multicolumn{1}{c}{$A_{\scalebox{0.8}{10}}$ ($\uparrow$)} & \multicolumn{1}{c}{$F_{\scalebox{0.8}{10}}$ ($\downarrow$)} & \multicolumn{1}{c}{$A_{\scalebox{0.8}{20}}$ ($\uparrow$)} & $F_{\scalebox{0.8}{20}}$ ($\downarrow$)
& \multicolumn{1}{c}{$A_{\scalebox{0.8}{10}}$ ($\uparrow$)} & \multicolumn{1}{c}{$F_{\scalebox{0.8}{10}}$ ($\downarrow$)} & \multicolumn{1}{c}{$A_{\scalebox{0.8}{20}}$ ($\uparrow$)} & $F_{\scalebox{0.8}{20}}$ ($\downarrow$) \\ \hline
Joint training             
& \multicolumn{1}{c}{79.60 $\pm$ 0.87} & \multicolumn{1}{c}{ -} & \multicolumn{1}{c}{79.60 $\pm$ 0.87} &  -    
& \multicolumn{1}{c}{93.22 $\pm$ 0.16} & \multicolumn{1}{c}{ -} & \multicolumn{1}{c}{93.22 $\pm$ 0.16} &  -    \\ \hline\hline
L2P~\cite{wang2022learning}          
& \multicolumn{1}{c}{63.49 $\pm$ 0.40} & \multicolumn{1}{c}{6.85 $\pm$ 0.42} & \multicolumn{1}{c}{59.38 $\pm$ 0.50} & 5.89 $\pm$ 0.36 
& \multicolumn{1}{c}{82.76 $\pm$ 1.17} & \multicolumn{1}{c}{7.86 $\pm$ 0.39} & \multicolumn{1}{c}{77.95 $\pm$ 0.72} & 9.88 $\pm$ 0.30  \\ 
DualPrompt~\cite{wang2022dualprompt}
& \multicolumn{1}{c}{68.50 $\pm$ 0.52} & \multicolumn{1}{c}{5.14 $\pm$ 0.18} & \multicolumn{1}{c}{63.21 $\pm$ 0.49} & 5.28 $\pm$ 0.45 
& \multicolumn{1}{c}{85.07 $\pm$ 0.49} & \multicolumn{1}{c}{5.57 $\pm$ 0.20} & \multicolumn{1}{c}{80.49 $\pm$ 0.31} & 8.84 $\pm$ 0.68 \\ 
DualPrompt-PGP \cite{qiao2023prompt}         
& \multicolumn{1}{c}{69.34 $\pm$ 0.05} & \multicolumn{1}{c}{4.53 $\pm$ 0.04} & \multicolumn{1}{c}{64.75 $\pm$ 0.38} & 6.04 $\pm$ 0.15 
& \multicolumn{1}{c}{86.92 $\pm$ 0.05} & \multicolumn{1}{c}{5.35 $\pm$ 0.19} & \multicolumn{1}{c}{83.74 $\pm$ 0.01} & 7.91 $\pm$ 0.15  \\ 
CODA-Prompt~\cite{smith2023coda}
& \multicolumn{1}{c}{74.24 $\pm$ 0.56} & \multicolumn{1}{c}{4.92 $\pm$ 0.21} & \multicolumn{1}{c}{70.86 $\pm$ 0.42} & 6.87 $\pm$ 0.25 
& \multicolumn{1}{c}{87.00 $\pm$ 0.38} & \multicolumn{1}{c}{4.78 $\pm$ 0.24} & \multicolumn{1}{c}{82.15 $\pm$ 0.17} & 6.33 $\pm$ 0.23    \\ 
CPrompt \cite{gao2024consistent} 
& \multicolumn{1}{c}{76.71 $\pm$ 0.61} & \multicolumn{1}{c}{4.66 $\pm$ 0.93} & \multicolumn{1}{c}{74.45 $\pm$ 0.25} & 4.98 $\pm$ 0.21
& \multicolumn{1}{c}{87.83 $\pm$ 0.37} & \multicolumn{1}{c}{4.88 $\pm$ 0.79} & \multicolumn{1}{c}{84.66 $\pm$ 0.13} & 5.69 $\pm$ 0.06  \\ 
ConvPrompt~\cite{roy2024convolutional} 
& \multicolumn{1}{c}{77.86 $\pm$ 0.25} & \multicolumn{1}{c}{4.33 $\pm$ 0.24} & \multicolumn{1}{c}{75.10 $\pm$ 0.39} & 4.10 $\pm$ 0.29 
& \multicolumn{1}{c}{88.87 $\pm$ 0.33} & \multicolumn{1}{c}{4.75 $\pm$ 0.15} & \multicolumn{1}{c}{87.37 $\pm$ 0.13} & 5.16 $\pm$ 0.01 \\ \hline\hline
\rowcolor[gray]{0.9} RainbowPrompt (ours)    
& \multicolumn{1}{c}{\textbf{79.09 $\pm$ 0.13}} & \multicolumn{1}{c}{\textbf{3.90 $\pm$ 0.23}} & \multicolumn{1}{c}{\textbf{78.36 $\pm$ 0.47}} & \textbf{3.44 $\pm$ 0.29} 
& \multicolumn{1}{c}{\textbf{89.86 $\pm$ 0.11}} & \multicolumn{1}{c}{\textbf{3.44 $\pm$ 0.26}} & \multicolumn{1}{c}{\textbf{90.15 $\pm$ 0.05}} & \textbf{3.75 $\pm$ 0.23} \\ \Xhline{1.pt} 
\end{tabular}}
\end{table*}

\section{Experiments}
\subsection{Setup}
\noindent{\textbf{Scenarios and Datasets.}}
To evaluate RainbowPrompt\footnote{We also represent RainbowPrompt as our method.}, we demonstrated it on image classification and video action recognition tasks under class-incremental learning scenarios.
For image classification, we used the ImageNet-R~\cite{hendrycks2021many}, CIFAR-100~\cite{krizhevsky2009learning}, and CUBS~\cite{wah2011caltech} datasets.
ImageNet-R is a challenging benchmark due to its classes featuring distinct styles (e.g., cartoon, graffiti, origami) and significant intra-class diversity~\cite{wang2022dualprompt}. 
We divided its 200 classes into disjoint subsets, forming 10-task (20 classes per task) and 20-task (10 classes per task) settings.
CIFAR-100, a widely used benchmark in prompt-based continual learning (PCL), was split into 10-task (10 classes per task) and 20-task (5 classes per task) settings.
CUBS, a fine-grained dataset featuring bird species, poses another challenge due to subtle inter-class differences. 
We applied the same task split configuration as used for ImageNet-R.
When splitting each dataset, we ensure no class overlaps occur between them.

For video action recognition, we conducted experiments using the vCLIMB benchmark~\cite{villa2022vclimb} on trimmed versions of the UCF-101~\cite{soomro2012ucf101} and ActivityNet~\cite{caba2015activitynet} datasets.
We evaluated our approach on UCF-101 (101 classes) and ActivityNet (200 classes) under both 10-task and 20-task settings. Following~\cite{villa2022vclimb}, we partitioned each dataset into disjoint subsets of randomly selected classes, with each subset treated as an independent task.

\noindent{\textbf{Compared Methods.}}
We evaluated RainbowPrompt against existing PCL approaches: L2P~\cite{wang2022learning}, DualPrompt~\cite{wang2022dualprompt}, DualPrompt-PGP \cite{qiao2023prompt}, CODA-Prompt~\cite{smith2023coda}, CPrompt \cite{gao2024consistent}, and ConvPrompt~\cite{roy2024convolutional}.
%We also reported the results of joint training, where the ViT model~\cite{dosovitskiy2020image} was trained on a combined dataset encompassing the training data from all tasks, serving as a baseline for comparison.
We also reported the results of joint training, where the ViT model~\cite{dosovitskiy2020image} was trained on a combined dataset encompassing the training data from all tasks.
We evaluated the methods using two metrics: average accuracy and average forgetting~\cite{wang2022learning}.

\noindent{\textbf{Implementation Details.}}
We followed the training details from \cite{smith2023coda}, including the optimizer and input resolution, while using a learning rate of 0.03.
We used a single base prompt with a prompt length of $L_p=20$ per task across all scenarios in RainbowPrompt.
We set the dimensionality of the projection matrices ($D_p$) to 96 for the 10-task setting and 56 for the 20-task setting.
We set the dimensionality of the weight matrices ($D_n$) as 56 for the 10-task setting and 28 for the 20-task setting.
$\text{LT}(\cdot)$ consists of two linear layers stacked with ReLU activations.
For adaptive prompting, we used soft decisions from the distribution $\delta_{l}^{t}$ during training for a certain number of epochs. 
After that, we sampled task-specific decisions from the learned distribution.
We implemented all methods, including RainbowPrompt, using the pre-trained ViT-B/16 as in~\cite{wang2022learning}.
We conducted each experiment three times with random trials and reported the results as the mean and standard deviation of the runs.

\begin{figure*}[t]
\includegraphics[width=\textwidth]{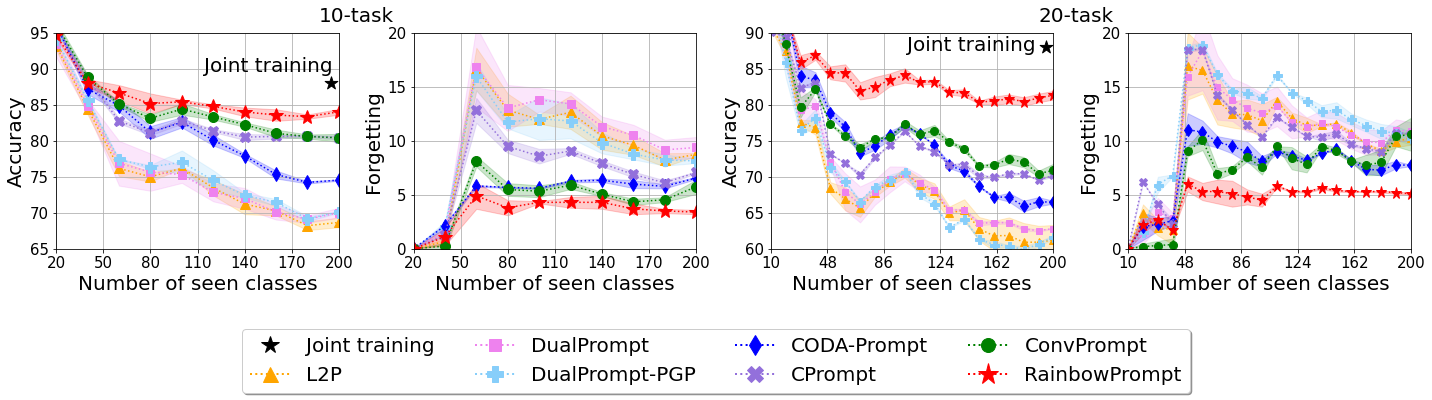}
%\vspace*{-5mm}
\caption{
Results on CUBS.
The average accuracy for all seen classes at each time step is shown with its standard deviation indicated by the shaded area.}
\label{fig:cubs}
\end{figure*}

\subsection{Image Classification}
We first evaluated the proposed approach on image classification using the ImageNet-R dataset. 
Tab. \ref{fig:combined} (left) presents the results for both 10-task and 20-task settings. 
%For ImageNet-R, RainbowPrompt achieves the highest average accuracy and the lowest average forgetting among all methods. 
In the 10-task setting, RainbowPrompt surpasses all the competitors, L2P, DualPrompt, DualPrompt-PGP, CODA-Prompt, CPrompt, and ConvPrompt, in average accuracy by gaps of 15.60\%, 10.59\%, 9.75\%, 4.85\%, 2.38\%, and 1.23\%, respectively. 
Compared to ConvPrompt, which achieves the lowest average forgetting among the compared methods, RainbowPrompt further reduces forgetting by 0.43\%, highlighting its superior capability to retain knowledge while accommodating new tasks.
The results remain consistent in the more challenging 20-task setting.
Notably, RainbowPrompt achieves an accuracy improvement of 3.26\% and a forgetting reduction of 0.66\% compared to the strong competitor, ConvPrompt.
These results highlight the effectiveness of ours in both the 10-task and 20-task settings.

\begin{table*}[t]
\caption{Results on the video action recognition tasks using UCF-101 and ActivityNet.}
\label{fig:video}
\centering
\resizebox{\linewidth}{!}{%
\begin{tabular}{l||cccc||cccc}
\Xhline{1.pt} 
\multirow{2}{*}{\textbf{Method}} 
& \multicolumn{4}{c}{\textbf{UCF-101}} & \multicolumn{4}{c}{\textbf{ActivityNet}} \\ \cline{2-9} 
& \multicolumn{1}{c}{$A_{\scalebox{0.8}{10}}$ ($\uparrow$)} & \multicolumn{1}{c}{$F_{\scalebox{0.8}{10}}$ ($\downarrow$)} & \multicolumn{1}{c}{$A_{\scalebox{0.8}{20}}$ ($\uparrow$)} & $F_{\scalebox{0.8}{20}}$ ($\downarrow$)
& \multicolumn{1}{c}{$A_{\scalebox{0.8}{10}}$ ($\uparrow$)} & \multicolumn{1}{c}{$F_{\scalebox{0.8}{10}}$ ($\downarrow$)} & \multicolumn{1}{c}{$A_{\scalebox{0.8}{20}}$ ($\uparrow$)} & $F_{\scalebox{0.8}{20}}$ ($\downarrow$) \\ \hline
L2P~\cite{wang2022learning}           
& \multicolumn{1}{c}{78.35 $\pm$ 1.54} & \multicolumn{1}{c}{5.46 $\pm$ 0.29} & \multicolumn{1}{c}{70.29 $\pm$ 0.19} & 8.31 $\pm$ 0.40 
& \multicolumn{1}{c}{63.46 $\pm$ 0.47} & \multicolumn{1}{c}{6.47 $\pm$ 1.13} & \multicolumn{1}{c}{56.47 $\pm$ 1.12} & 9.96 $\pm$ 1.33  \\ 
DualPrompt~\cite{wang2022dualprompt}    
& \multicolumn{1}{c}{83.15 $\pm$ 0.82} & \multicolumn{1}{c}{5.41 $\pm$ 0.38} & \multicolumn{1}{c}{74.96 $\pm$ 1.23} & 7.97 $\pm$ 0.41 
& \multicolumn{1}{c}{63.76 $\pm$ 0.88} & \multicolumn{1}{c}{5.78 $\pm$ 0.50} & \multicolumn{1}{c}{56.59 $\pm$ 0.81} & 9.89 $\pm$ 0.05 \\ 
DualPrompt-PGP \cite{qiao2023prompt}          
& \multicolumn{1}{c}{84.39 $\pm$ 0.56} & \multicolumn{1}{c}{5.92 $\pm$ 0.16} & \multicolumn{1}{c}{75.37 $\pm$ 0.16} & 7.50 $\pm$ 0.59 
& \multicolumn{1}{c}{64.65 $\pm$ 0.36} & \multicolumn{1}{c}{6.20 $\pm$ 0.64} & \multicolumn{1}{c}{57.46 $\pm$ 0.64} & 8.56 $\pm$ 0.25  \\ 
CODA-Prompt~\cite{smith2023coda}     
& \multicolumn{1}{c}{84.77 $\pm$ 0.75} & \multicolumn{1}{c}{5.81 $\pm$ 0.27} & \multicolumn{1}{c}{75.35 $\pm$ 0.90} & 5.66 $\pm$ 0.50 
& \multicolumn{1}{c}{66.13 $\pm$ 0.65} & \multicolumn{1}{c}{6.23 $\pm$ 0.15} & \multicolumn{1}{c}{58.62 $\pm$ 0.58} & 8.27 $\pm$ 0.34    \\ 
CPrompt \cite{gao2024consistent}         
& \multicolumn{1}{c}{87.16 $\pm$ 0.59} & \multicolumn{1}{c}{5.30 $\pm$ 0.25} & \multicolumn{1}{c}{81.78 $\pm$ 0.04} & 4.42 $\pm$ 0.19
& \multicolumn{1}{c}{66.81 $\pm$ 0.48} & \multicolumn{1}{c}{7.25 $\pm$ 0.08} & \multicolumn{1}{c}{62.17 $\pm$ 0.66} & 7.49 $\pm$ 0.57  \\ 
ConvPrompt~\cite{roy2024convolutional}       
& \multicolumn{1}{c}{85.58 $\pm$ 0.50} & \multicolumn{1}{c}{5.34 $\pm$ 0.25} & \multicolumn{1}{c}{78.83 $\pm$ 0.29} & 4.03 $\pm$ 0.79 
& \multicolumn{1}{c}{67.32 $\pm$ 0.30} & \multicolumn{1}{c}{5.04 $\pm$ 0.29} & \multicolumn{1}{c}{60.01 $\pm$ 0.34} & 6.02 $\pm$ 0.52 \\ \hline\hline
\rowcolor[gray]{0.9} RainbowPrompt (ours)    
& \multicolumn{1}{c}{\textbf{89.03 $\pm$ 0.91}} & \multicolumn{1}{c}{\textbf{4.91 $\pm$ 0.30}} & \multicolumn{1}{c}{\textbf{84.05 $\pm$ 1.27}} & \textbf{3.59 $\pm$ 0.17} 
& \multicolumn{1}{c}{\textbf{69.87 $\pm$ 0.24}} & \multicolumn{1}{c}{\textbf{3.96 $\pm$ 0.28}} & \multicolumn{1}{c}{\textbf{70.55 $\pm$ 1.57}} & \textbf{5.47 $\pm$ 0.47} \\ \Xhline{1.pt} 
\end{tabular}}
\end{table*}

We further evaluated RainbowPrompt on CIFAR-100. 
The results for the 10-task and 20-task settings are summarized in Tab. \ref{fig:combined} (right). 
%RainbowPrompt demonstrates outstanding performance in both settings.
Notably, in the 20-task setting, L2P, DualPrompt, DualPrompt-PGP, CODA-Prompt, CPrompt, and ConvPrompt exhibit accuracy drops of 4.81\%, 4.58\%, 3.18\%, 4.85\%, 3.17\%, and 1.50\%, respectively, compared to their performance in the 10-task setting. 
In contrast, RainbowPrompt not only avoids performance degradation but achieves a 0.29\% increase in accuracy on average. 
It demonstrates consistently competitive performance as the number of tasks increases, while others struggle to adapt under extended task sequences.

We evaluated RainbowPrompt on the fine-grained CUBS dataset. 
Fig. \ref{fig:cubs} illustrates the average accuracy and forgetting across all seen classes. 
RainbowPrompt outperforms all baseline methods in terms of the metrics.
In the 10-task setting, RainbowPrompt and the strongest baseline, ConvPrompt, exhibit stable performance as the number of seen classes increases. 
RainbowPrompt achieves a 3.62\% higher accuracy and 2.28\% lower forgetting than ConvPrompt.
Notably, in the 20-task setting, RainbowPrompt significantly outperforms ConvPrompt, achieving a notable 10.44\% improvement in average accuracy and a 5.44\% reduction in forgetting.
It demonstrates the ability to produce diversity-enhanced prompts while accommodating new tasks.

\subsection{Video Action Recognition}
To extend the evaluation of the proposed prompt-evolving mechanism beyond image classification, we applied it to video action recognition tasks.
We utilized trimmed versions of the UCF-101 and ActivityNet datasets \cite{villa2023pivot} to focus on well-defined actions and facilitate analysis by excluding irrelevant background segments.
Each video was divided into three equal segments, and one frame was randomly sampled from each segment~\cite{wang2018temporal}. 
The datasets were prepared using the temporal segment sampling strategy outlined in~\cite{wang2018temporal}.
We evaluated the proposed approach on the 10-task and 20-task settings for each dataset, with the results presented in Tab. \ref{fig:video}.

\begin{figure*}[t]
\includegraphics[width=0.96\linewidth]{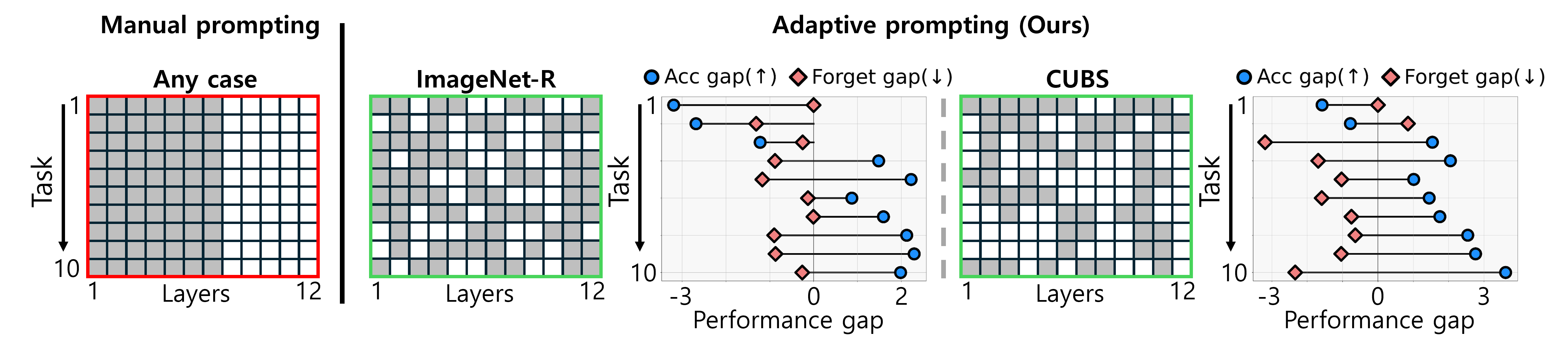}
%\vspace*{-1.5mm}
\caption{
Visualization of prompt insertion patterns across layers: manual prompting (red box)~\cite{roy2024convolutional} vs. the proposed adaptive prompting (green boxes).
Gray represents layers with inserted prompts, whereas white denotes layers without prompts.
Accuracy and forgetting gaps denote the average differences in performance across all seen classes, measured as the difference between ours and manual prompting.
}
\label{fig:adpative_cubs}
\end{figure*}

In the 10-task setting on UCF-101, RainbowPrompt outperforms L2P, DualPrompt, DualPrompt-PGP, CODA-Prompt, CPrompt, and ConvPrompt in average accuracy by 10.68\%, 5.88\%, 4.64\%, 4.26\%, 1.87\%, and 3.45\%, respectively. 
Also, it reduces forgetting by 0.55\%, 0.50\%, 1.01\%, 0.90\%, 0.39\%, and 0.43\% compared to these methods.
%In the 20-task setting, it outperforms all competing methods, achieving an average improvement of 7.95\% in accuracy and a 3.92\% reduction in forgetting.
%In particular, when compared to the most competitive methods, CPrompt and ConvPrompt, RainbowPrompt achieves an average improvement of 3.74\% in accuracy and a 0.63\% reduction in forgetting.
%This demonstrates the effectiveness of the proposed method in handling tasks involving temporal information.
In the 20-task setting, it outperforms all competing methods by a larger gap than in the 10-task setting.
Specifically, it achieves an average improvement of 7.95\% in accuracy and a 3.92\% reduction in forgetting compared to other methods.
This is because a larger number of tasks enhances diversity of accumulated knowledge, allowing for a more effective evolution process.

In the 10-task setting on ActivityNet, RainbowPrompt outperforms CPrompt and ConvPrompt by achieving 3.06\% higher accuracy and 3.29\% lower forgetting than CPrompt and an additional 2.55\% accuracy gain and 1.08\% lower forgetting over ConvPrompt.
%In the 20-task setting, it achieves a 8.38\% improvement in accuracy and a 2.02\% reduction in forgetting compared to CPrompt and surpasses ConvPrompt with a 10.54\% accuracy gain and a 0.55\% reduction in forgetting.
%Unlike them, the proposed method yields more discriminative predictions~\cite{cui2020towards}, enabling RainbowPrompt to handle complex action classes more effectively.
In the 20-task setting, it achieves a significant performance gain over ConvPrompt, with a 10.54\% improvement in accuracy and a 0.55\% reduction in forgetting. 
Notably, while ConvPrompt experiences a 7.31\% drop in accuracy from its 10-task setting, RainbowPrompt not only prevents degradation but instead improves by 0.68\%. 
The diversity-enhancing integration of RainbowPrompt promotes predictive discriminability~\cite{cui2020towards}, enabling it to handle complex action classes more effectively.

\begin{table}[t]
\centering
\caption{
Ablation study of RainbowPrompt on ImageNet-R.}
\label{fig:ablation}
\renewcommand{\arraystretch}{1.17}
\setlength{\tabcolsep}{9pt}
\resizebox{\columnwidth}{!}{%
\begin{tabular}{ccccc||cc}
\Xhline{1.pt}
\centering
TC & TLT & FLT & TGA & AP & $A_{\scalebox{0.8}{10}}$ ($\uparrow$) & $F_{\scalebox{0.8}{10}}$ ($\downarrow$) \\ \hline\hline
\rowcolor[gray]{0.9} \checkmark  &  \checkmark   &  \checkmark   &  \checkmark   & \checkmark   & \textbf{79.09 $\pm$ 0.13}  & \textbf{3.90 $\pm$ 0.23} \\ \hline
 -  &  \checkmark   &  \checkmark   &  \checkmark   &  \checkmark  & 78.92 $\pm$ 0.06  & 4.19 $\pm$ 0.14  \\ \hline
 \checkmark  &  -   &  \checkmark   &  \checkmark   &  \checkmark  & 78.70 $\pm$ 0.41  & 4.14 $\pm$ 0.00  \\ \hline
 \checkmark  &  \checkmark   & -    &  \checkmark   &  \checkmark  & 78.57 $\pm$ 0.11  & 4.29 $\pm$ 0.36  \\ \hline
 \checkmark  &  \checkmark   &  \checkmark   &  -   & \checkmark   & 66.31 $\pm$ 0.48  & 4.84 $\pm$ 0.26  \\ \hline
 \checkmark  &  \checkmark   &  \checkmark   &  \checkmark   &  -  & 78.13 $\pm$ 0.22  & 4.07 $\pm$ 0.19  \\ \Xhline{1.pt} 
\end{tabular}}
\end{table}

\subsection{Analysis}
\noindent{\textbf{Ablation Study.}}
We conducted an ablation study on ImageNet-R in the 10-task setting to evaluate the impacts of the key components in RainbowPrompt.
Tab. \ref{fig:ablation} presents the effect of excluding each component, including the task-conditioning step (TC), task-level transformation (TLT), feature-level transformation (FLT), task-guided alignment (TGA), and adaptive prompting (AP).
Excluding the task-conditioning step results in small declines in average accuracy and forgetting, suggesting that it provides supplementary task-specific information.
%Between task-level and feature-level transformations, removing the feature-level transformation leads to a greater performance drop, as it captures fine-grained element-level information in each prompt.
Compared to the task-level transformation, which captures overall task dependencies, the feature-level transformation plays a more critical role by capturing fine-grained feature interactions, leading to a larger performance drop when removed.
Excluding the task-guided alignment leads to the most significant decline in performance, underscoring its essential role in RainbowPrompt. 
It aligns prompts based on their transformed representations, creating a coherent prompt that facilitates new task learning and integrates diverse knowledge.
We also observe that replacing manual prompting with an adaptive approach yields performance gains through task-specific tuning strategies.

\noindent{\textbf{Adaptive Prompting.}}
\label{sec: adaptive}
We conducted an empirical analysis to demonstrate the advantages of the proposed adaptive prompting, as shown in Fig. \ref{fig:adpative_cubs}.
We visualize the task-specific prompt insertion layers of manual~\cite{roy2024convolutional} and adaptive prompting in the 10-task setting using ImageNet-R and CUBS.
Manual prompting applies predetermined prompt-insertion layers uniformly across all tasks \cite{smith2023coda, gao2024consistent}, regardless of the dataset.
In contrast, our adaptive prompting dynamically determines the optimal prompt-insertion layers for each dataset and task.
For ImageNet-R, adaptive prompting outperforms manual prompting, achieving a 1.98\% accuracy gain and 0.26\% less forgetting across all tasks.
We observe a drop in accuracy in the first three tasks, while forgetting remains consistently lower or comparable to manual prompting.
Note here that integrating historical knowledge is crucial for transferring knowledge to new tasks, but its absence in early sequential learning limits accuracy as shown in the figure.
For CUBS, adaptive prompting achieves a 3.61\% higher average accuracy and 2.34\% lower forgetting than manual prompting.
Our approach demonstrates exceptional performance across most tasks, with even larger performance gaps than those observed on ImageNet-R.

\begin{figure}[t]
    \centering
    \includegraphics[width=1.\linewidth]{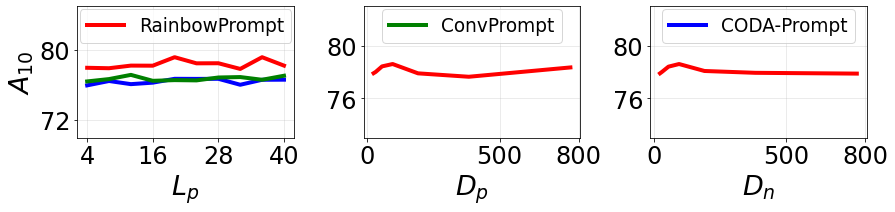}
    %\vspace{-6mm}
    \caption{Sensitivity analysis on ImageNet-R.}
    \label{fig:sa}
\end{figure}

\noindent{\textbf{Sensitivity.}}
To assess the robustness of our method to hyper-parameter variations, we performed sensitivity analysis on the prompt length ($L_p$) and the dimensions of the projection and weight matrices ($D_p$, $D_n$). 
As illustrated in Fig. \ref{fig:sa}, RainbowPrompt consistently outperforms the strong baselines, CODA-Prompt and ConvPrompt, across various $L_p$ values (see first figure). 
Additionally, varying $D_p$ or $D_n$ shows low sensitivity to changes in dimensionality, as shown in the second and third figures.
These results indicate that our method is robust to hyper-parameter choices and maintains stable performance.

\noindent{\textbf{Efficiency.}}
To evaluate the computational efficiency of our approach, we assessed performance on CIFAR-100 (20-task).
CODA-Prompt and ConvPrompt incur 33.7B and 17.1B MACs with 4.8M and 5.7M trainable parameters.
%Our method trains with 8.2M parameters but discards 76.5\% (6.2M) at inference, requiring only 18.5B MACs.
Our method trains with 8.2M parameters, including prompt-evolving components, but discards 76.5\% (6.2M) at inference, requiring only 18.5B MACs.
This offers a favorable balance between accuracy and efficiency.

\section{Conclusion}
We have proposed RainbowPrompt, a novel prompt-evolving mechanism to address a fundamental challenge in sequential task learning, which effectively integrates accumulated historical knowledge without compromising previously learned information.
Unlike existing methods, RainbowPrompt progressively improves the representations of task-specific prompts for new tasks to promote integration that enhances diversity.
By dynamically regulating layer activation with a learnable probabilistic gate, our method optimizes the evolution process based on task-specific differences.
Extensive experiments on image classification and video action recognition benchmarks demonstrate that RainbowPrompt consistently outperforms state-of-the-art methods, achieving significant improvements in accuracy and forgetting across diverse scenarios.

\newpage
\noindent{\textbf{Acknowledgements.}} This work was supported in part by the National Research Foundation of Korea (NRF) grant funded by the Korea government(MSIT) (RS-2023-00279019) and in part by the Institute of Information \& Communications Technology Planning \& Evaluation (IITP) grant funded by the Korea government (MSIT) [RS-2021-II211341, Artificial Intelligence Graduate School Program (Chung-Ang University)].

{
    
    \bibliographystyle{ieeenat_fullname}
    \bibliography{main}
}


\end{document}

%% file: main.bbl
\begin{thebibliography}{48}
\providecommand{\natexlab}[1]{#1}
\providecommand{\url}[1]{\texttt{#1}}
\expandafter\ifx\csname urlstyle\endcsname\relax
  \providecommand{\doi}[1]{doi: #1}\else
  \providecommand{\doi}{doi: \begingroup \urlstyle{rm}\Url}\fi

\bibitem[Aljundi et~al.(2018)Aljundi, Babiloni, Elhoseiny, Rohrbach, and Tuytelaars]{aljundi2018memory}
Rahaf Aljundi, Francesca Babiloni, Mohamed Elhoseiny, Marcus Rohrbach, and Tinne Tuytelaars.
\newblock Memory aware synapses: Learning what (not) to forget.
\newblock In \emph{Proceedings of the European Conference on Computer Vision (ECCV)}, pages 139--154, 2018.

\bibitem[Caba~Heilbron et~al.(2015)Caba~Heilbron, Escorcia, Ghanem, and Carlos~Niebles]{caba2015activitynet}
Fabian Caba~Heilbron, Victor Escorcia, Bernard Ghanem, and Juan Carlos~Niebles.
\newblock Activitynet: A large-scale video benchmark for human activity understanding.
\newblock In \emph{Proceedings of the ieee conference on computer vision and pattern recognition}, pages 961--970, 2015.

\bibitem[Cha et~al.(2021)Cha, Lee, and Shin]{cha2021co2l}
Hyuntak Cha, Jaeho Lee, and Jinwoo Shin.
\newblock Co2l: Contrastive continual learning.
\newblock In \emph{Proceedings of the IEEE/CVF International conference on computer vision}, pages 9516--9525, 2021.

\bibitem[Chaudhry et~al.(2018)Chaudhry, Ranzato, Rohrbach, and Elhoseiny]{chaudhry2018efficient}
Arslan Chaudhry, Marc'Aurelio Ranzato, Marcus Rohrbach, and Mohamed Elhoseiny.
\newblock Efficient lifelong learning with a-gem.
\newblock \emph{arXiv preprint arXiv:1812.00420}, 2018.

\bibitem[Chen and Liu(2018)]{chen2018lifelong}
Zhiyuan Chen and Bing Liu.
\newblock \emph{Lifelong machine learning}.
\newblock Morgan \& Claypool Publishers, 2018.

\bibitem[Cui et~al.(2020)Cui, Wang, Zhuo, Li, Huang, and Tian]{cui2020towards}
Shuhao Cui, Shuhui Wang, Junbao Zhuo, Liang Li, Qingming Huang, and Qi Tian.
\newblock Towards discriminability and diversity: Batch nuclear-norm maximization under label insufficient situations.
\newblock In \emph{Proceedings of the IEEE/CVF conference on computer vision and pattern recognition}, pages 3941--3950, 2020.

\bibitem[De~Lange et~al.(2021)De~Lange, Aljundi, Masana, Parisot, Jia, Leonardis, Slabaugh, and Tuytelaars]{de2021continual}
Matthias De~Lange, Rahaf Aljundi, Marc Masana, Sarah Parisot, Xu Jia, Ale{\v{s}} Leonardis, Gregory Slabaugh, and Tinne Tuytelaars.
\newblock A continual learning survey: Defying forgetting in classification tasks.
\newblock \emph{IEEE Transactions on Pattern Analysis and Machine Intelligence}, 44\penalty0 (7):\penalty0 3366--3385, 2021.

\bibitem[Dosovitskiy(2020)]{dosovitskiy2020image}
Alexey Dosovitskiy.
\newblock An image is worth 16x16 words: Transformers for image recognition at scale.
\newblock \emph{arXiv preprint arXiv:2010.11929}, 2020.

\bibitem[French(1999)]{french1999catastrophic}
Robert~M French.
\newblock Catastrophic forgetting in connectionist networks.
\newblock \emph{Trends in Cognitive Sciences}, 3\penalty0 (4):\penalty0 128--135, 1999.

\bibitem[Gao et~al.(2016)Gao, Beijbom, Zhang, and Darrell]{gao2016compact}
Yang Gao, Oscar Beijbom, Ning Zhang, and Trevor Darrell.
\newblock Compact bilinear pooling.
\newblock In \emph{Proceedings of the IEEE conference on computer vision and pattern recognition}, pages 317--326, 2016.

\bibitem[Gao et~al.(2024)Gao, Cen, and Chang]{gao2024consistent}
Zhanxin Gao, Jun Cen, and Xiaobin Chang.
\newblock Consistent prompting for rehearsal-free continual learning.
\newblock In \emph{Proceedings of the IEEE/CVF Conference on Computer Vision and Pattern Recognition}, pages 28463--28473, 2024.

\bibitem[Geva et~al.(2020)Geva, Schuster, Berant, and Levy]{geva2020transformer}
Mor Geva, Roei Schuster, Jonathan Berant, and Omer Levy.
\newblock Transformer feed-forward layers are key-value memories.
\newblock \emph{arXiv preprint arXiv:2012.14913}, 2020.

\bibitem[Han et~al.(2021)Han, Yun, Heo, and Yoo]{han2021rethinking}
Dongyoon Han, Sangdoo Yun, Byeongho Heo, and YoungJoon Yoo.
\newblock Rethinking channel dimensions for efficient model design.
\newblock In \emph{Proceedings of the IEEE/CVF conference on Computer Vision and Pattern Recognition}, pages 732--741, 2021.

\bibitem[Hendrycks et~al.(2021)Hendrycks, Basart, Mu, Kadavath, Wang, Dorundo, Desai, Zhu, Parajuli, Guo, et~al.]{hendrycks2021many}
Dan Hendrycks, Steven Basart, Norman Mu, Saurav Kadavath, Frank Wang, Evan Dorundo, Rahul Desai, Tyler Zhu, Samyak Parajuli, Mike Guo, et~al.
\newblock The many faces of robustness: A critical analysis of out-of-distribution generalization.
\newblock In \emph{Proceedings of the IEEE/CVF international conference on computer vision}, pages 8340--8349, 2021.

\bibitem[Hong et~al.(2025)Hong, Jin, Suh, and Kim]{hong2025exploration}
Kiseong Hong, Hyundong Jin, Sungho Suh, and Eunwoo Kim.
\newblock Exploration and exploitation in continual learning.
\newblock \emph{Neural Networks}, 188:\penalty0 107444, 2025.

\bibitem[Hsu(2018)]{hsu2018re}
Y Hsu.
\newblock Re-evaluating continual learning scenarios: A categorization and case for strong baselines.
\newblock \emph{arXiv preprint arXiv:1810.12488}, 2018.

\bibitem[Hung et~al.(2019)Hung, Tu, Wu, Chen, Chan, and Chen]{hung2019compacting}
Ching-Yi Hung, Cheng-Hao Tu, Cheng-En Wu, Chien-Hung Chen, Yi-Ming Chan, and Chu-Song Chen.
\newblock Compacting, picking and growing for unforgetting continual learning.
\newblock \emph{Advances in Neural Information Processing Systems}, 32, 2019.

\bibitem[Jang et~al.(2016)Jang, Gu, and Poole]{jang2016categorical}
Eric Jang, Shixiang Gu, and Ben Poole.
\newblock Categorical reparameterization with gumbel-softmax.
\newblock \emph{arXiv preprint arXiv:1611.01144}, 2016.

\bibitem[Jin and Kim(2022)]{jin2022helpful}
Hyundong Jin and Eunwoo Kim.
\newblock Helpful or harmful: Inter-task association in continual learning.
\newblock In \emph{European Conference on Computer Vision}, pages 519--535. Springer, 2022.

\bibitem[Jin et~al.(2023)Jin, Kim, Ahn, and Kim]{jin2023growing}
Hyundong Jin, Gyeong-hyeon Kim, Chanho Ahn, and Eunwoo Kim.
\newblock Growing a brain with sparsity-inducing generation for continual learning.
\newblock In \emph{Proceedings of the IEEE/CVF international conference on computer vision}, pages 18961--18970, 2023.

\bibitem[Jung et~al.(2023)Jung, Han, Bang, and Song]{jung2023generating}
Dahuin Jung, Dongyoon Han, Jihwan Bang, and Hwanjun Song.
\newblock Generating instance-level prompts for rehearsal-free continual learning.
\newblock In \emph{Proceedings of the IEEE/CVF International Conference on Computer Vision}, pages 11847--11857, 2023.

\bibitem[Jung et~al.(2020)Jung, Ahn, Cha, and Moon]{jung2020continual}
Sangwon Jung, Hongjoon Ahn, Sungmin Cha, and Taesup Moon.
\newblock Continual learning with node-importance based adaptive group sparse regularization.
\newblock \emph{Advances in Neural Information Processing Systems}, 33:\penalty0 3647--3658, 2020.

\bibitem[Khattak et~al.(2023)Khattak, Wasim, Naseer, Khan, Yang, and Khan]{khattak2023self}
Muhammad~Uzair Khattak, Syed~Talal Wasim, Muzammal Naseer, Salman Khan, Ming-Hsuan Yang, and Fahad~Shahbaz Khan.
\newblock Self-regulating prompts: Foundational model adaptation without forgetting.
\newblock In \emph{Proceedings of the IEEE/CVF International Conference on Computer Vision}, pages 15190--15200, 2023.

\bibitem[Krizhevsky et~al.(2009)Krizhevsky, Hinton, et~al.]{krizhevsky2009learning}
Alex Krizhevsky, Geoffrey Hinton, et~al.
\newblock Learning multiple layers of features from tiny images.
\newblock 2009.

\bibitem[Lester et~al.(2021)Lester, Al-Rfou, and Constant]{lester2021power}
Brian Lester, Rami Al-Rfou, and Noah Constant.
\newblock The power of scale for parameter-efficient prompt tuning.
\newblock \emph{arXiv preprint arXiv:2104.08691}, 2021.

\bibitem[Li and Liang(2021)]{li2021prefix}
Xiang~Lisa Li and Percy Liang.
\newblock Prefix-tuning: Optimizing continuous prompts for generation.
\newblock \emph{arXiv preprint arXiv:2101.00190}, 2021.

\bibitem[Liang and Li(2024)]{liang2024inflora}
Yan-Shuo Liang and Wu-Jun Li.
\newblock Inflora: Interference-free low-rank adaptation for continual learning.
\newblock In \emph{Proceedings of the IEEE/CVF Conference on Computer Vision and Pattern Recognition}, pages 23638--23647, 2024.

\bibitem[Liu et~al.(2023)Liu, Yuan, Fu, Jiang, Hayashi, and Neubig]{liu2023pre}
Pengfei Liu, Weizhe Yuan, Jinlan Fu, Zhengbao Jiang, Hiroaki Hayashi, and Graham Neubig.
\newblock Pre-train, prompt, and predict: A systematic survey of prompting methods in natural language processing.
\newblock \emph{ACM Computing Surveys}, 55\penalty0 (9):\penalty0 1--35, 2023.

\bibitem[Lopez-Paz and Ranzato(2017)]{lopez2017gradient}
David Lopez-Paz and Marc'Aurelio Ranzato.
\newblock Gradient episodic memory for continual learning.
\newblock \emph{Advances in Neural Information Processing Systems}, 30, 2017.

\bibitem[Ma et~al.(2023)Ma, Liu, Deng, Xie, Dong, and Xu]{ma2023understanding}
Chengcheng Ma, Yang Liu, Jiankang Deng, Lingxi Xie, Weiming Dong, and Changsheng Xu.
\newblock Understanding and mitigating overfitting in prompt tuning for vision-language models.
\newblock \emph{IEEE Transactions on Circuits and Systems for Video Technology}, 33\penalty0 (9):\penalty0 4616--4629, 2023.

\bibitem[Meng et~al.(2022)Meng, Bau, Andonian, and Belinkov]{meng2022locating}
Kevin Meng, David Bau, Alex Andonian, and Yonatan Belinkov.
\newblock Locating and editing factual associations in gpt.
\newblock \emph{Advances in Neural Information Processing Systems}, 35:\penalty0 17359--17372, 2022.

\bibitem[Ostapenko et~al.(2019)Ostapenko, Puscas, Klein, Jahnichen, and Nabi]{ostapenko2019learning}
Oleksiy Ostapenko, Mihai Puscas, Tassilo Klein, Patrick Jahnichen, and Moin Nabi.
\newblock Learning to remember: A synaptic plasticity driven framework for continual learning.
\newblock In \emph{Proceedings of the IEEE/CVF Conference on Computer Vision and Pattern Recognition}, pages 11321--11329, 2019.

\bibitem[Qiao et~al.(2023)Qiao, Tan, Chen, Qu, Peng, Xie, et~al.]{qiao2023prompt}
Jingyang Qiao, Xin Tan, Chengwei Chen, Yanyun Qu, Yong Peng, Yuan Xie, et~al.
\newblock Prompt gradient projection for continual learning.
\newblock In \emph{The Twelfth International Conference on Learning Representations}, 2023.

\bibitem[Rebuffi et~al.(2017)Rebuffi, Kolesnikov, Sperl, and Lampert]{rebuffi2017icarl}
Sylvestre-Alvise Rebuffi, Alexander Kolesnikov, Georg Sperl, and Christoph~H Lampert.
\newblock icarl: Incremental classifier and representation learning.
\newblock In \emph{Proceedings of the IEEE Conference on Computer Vision and Pattern Recognition}, pages 2001--2010, 2017.

\bibitem[Roy et~al.(2024)Roy, Moulick, Verma, Ghosh, and Das]{roy2024convolutional}
Anurag Roy, Riddhiman Moulick, Vinay~K Verma, Saptarshi Ghosh, and Abir Das.
\newblock Convolutional prompting meets language models for continual learning.
\newblock In \emph{Proceedings of the IEEE/CVF Conference on Computer Vision and Pattern Recognition}, pages 23616--23626, 2024.

\bibitem[Saha et~al.(2021)Saha, Garg, and Roy]{saha2021gradient}
Gobinda Saha, Isha Garg, and Kaushik Roy.
\newblock Gradient projection memory for continual learning.
\newblock \emph{arXiv preprint arXiv:2103.09762}, 2021.

\bibitem[Shin et~al.(2017)Shin, Lee, Kim, and Kim]{shin2017continual}
Hanul Shin, Jung~Kwon Lee, Jaehong Kim, and Jiwon Kim.
\newblock Continual learning with deep generative replay.
\newblock \emph{Advances in Neural Information Processing Systems}, 30, 2017.

\bibitem[Smith et~al.(2023)Smith, Karlinsky, Gutta, Cascante-Bonilla, Kim, Arbelle, Panda, Feris, and Kira]{smith2023coda}
James~Seale Smith, Leonid Karlinsky, Vyshnavi Gutta, Paola Cascante-Bonilla, Donghyun Kim, Assaf Arbelle, Rameswar Panda, Rogerio Feris, and Zsolt Kira.
\newblock Coda-prompt: Continual decomposed attention-based prompting for rehearsal-free continual learning.
\newblock In \emph{Proceedings of the IEEE/CVF Conference on Computer Vision and Pattern Recognition}, pages 11909--11919, 2023.

\bibitem[Soomro(2012)]{soomro2012ucf101}
K Soomro.
\newblock Ucf101: A dataset of 101 human actions classes from videos in the wild.
\newblock \emph{arXiv preprint arXiv:1212.0402}, 2012.

\bibitem[Villa et~al.(2022)Villa, Alhamoud, Escorcia, Caba, Alc{\'a}zar, and Ghanem]{villa2022vclimb}
Andr{\'e}s Villa, Kumail Alhamoud, Victor Escorcia, Fabian Caba, Juan~Le{\'o}n Alc{\'a}zar, and Bernard Ghanem.
\newblock vclimb: A novel video class incremental learning benchmark.
\newblock In \emph{Proceedings of the IEEE/CVF Conference on Computer Vision and Pattern Recognition}, pages 19035--19044, 2022.

\bibitem[Villa et~al.(2023)Villa, Alc{\'a}zar, Alfarra, Alhamoud, Hurtado, Heilbron, Soto, and Ghanem]{villa2023pivot}
Andr{\'e}s Villa, Juan~Le{\'o}n Alc{\'a}zar, Motasem Alfarra, Kumail Alhamoud, Julio Hurtado, Fabian~Caba Heilbron, Alvaro Soto, and Bernard Ghanem.
\newblock Pivot: Prompting for video continual learning.
\newblock In \emph{Proceedings of the IEEE/CVF Conference on Computer Vision and Pattern Recognition}, pages 24214--24223, 2023.

\bibitem[Wah et~al.(2011)Wah, Branson, Welinder, Perona, and Belongie]{wah2011caltech}
Catherine Wah, Steve Branson, Peter Welinder, Pietro Perona, and Serge Belongie.
\newblock The caltech-ucsd birds-200-2011 dataset.
\newblock 2011.

\bibitem[Wang et~al.(2018)Wang, Xiong, Wang, Qiao, Lin, Tang, and Van~Gool]{wang2018temporal}
Limin Wang, Yuanjun Xiong, Zhe Wang, Yu Qiao, Dahua Lin, Xiaoou Tang, and Luc Van~Gool.
\newblock Temporal segment networks for action recognition in videos.
\newblock \emph{IEEE transactions on pattern analysis and machine intelligence}, 41\penalty0 (11):\penalty0 2740--2755, 2018.

\bibitem[Wang et~al.(2022{\natexlab{a}})Wang, Zhang, Ebrahimi, Sun, Zhang, Lee, Ren, Su, Perot, Dy, et~al.]{wang2022dualprompt}
Zifeng Wang, Zizhao Zhang, Sayna Ebrahimi, Ruoxi Sun, Han Zhang, Chen-Yu Lee, Xiaoqi Ren, Guolong Su, Vincent Perot, Jennifer Dy, et~al.
\newblock Dualprompt: Complementary prompting for rehearsal-free continual learning.
\newblock In \emph{European Conference on Computer Vision}, pages 631--648. Springer, 2022{\natexlab{a}}.

\bibitem[Wang et~al.(2022{\natexlab{b}})Wang, Zhang, Lee, Zhang, Sun, Ren, Su, Perot, Dy, and Pfister]{wang2022learning}
Zifeng Wang, Zizhao Zhang, Chen-Yu Lee, Han Zhang, Ruoxi Sun, Xiaoqi Ren, Guolong Su, Vincent Perot, Jennifer Dy, and Tomas Pfister.
\newblock Learning to prompt for continual learning.
\newblock In \emph{Proceedings of the IEEE/CVF conference on computer vision and pattern recognition}, pages 139--149, 2022{\natexlab{b}}.

\bibitem[Wu et~al.(2018)Wu, Herranz, Liu, Van De~Weijer, Raducanu, et~al.]{wu2018memory}
Chenshen Wu, Luis Herranz, Xialei Liu, Joost Van De~Weijer, Bogdan Raducanu, et~al.
\newblock Memory replay gans: Learning to generate new categories without forgetting.
\newblock \emph{Advances in Neural Information Processing Systems}, 31, 2018.

\bibitem[Yan et~al.(2021)Yan, Xie, and He]{yan2021dynamically}
Shipeng Yan, Jiangwei Xie, and Xuming He.
\newblock Der: Dynamically expandable representation for class incremental learning.
\newblock In \emph{Proceedings of the IEEE/CVF Conference on Computer Vision and Pattern Recognition}, pages 3014--3023, 2021.

\bibitem[Yoon et~al.(2017)Yoon, Yang, Lee, and Hwang]{yoon2017lifelong}
Jaehong Yoon, Eunho Yang, Jeongtae Lee, and Sung~Ju Hwang.
\newblock Lifelong learning with dynamically expandable networks.
\newblock \emph{arXiv preprint arXiv:1708.01547}, 2017.

\end{thebibliography}
